\documentclass[10pt,conference,letterpaper]{IEEEtran}
\IEEEoverridecommandlockouts

\usepackage{amsmath,amssymb}
\usepackage{graphicx}
\graphicspath{{./}}
\usepackage{booktabs}
\usepackage{flushend}
\usepackage{tabularx}
\usepackage{array}
\usepackage{float}
\usepackage{etoolbox}
\usepackage{placeins}
\usepackage[font=small,labelfont=bf,labelsep=period,justification=raggedright,singlelinecheck=false]{caption}
\usepackage[table]{xcolor}
\usepackage{xurl}
\usepackage[american]{babel}
\usepackage{csquotes}
\usepackage[backend=biber,style=apa,natbib=true]{biblatex}
\usepackage[hidelinks]{hyperref}
\hypersetup{pdftitle={SPIRE: Task-agnostic embodied continual learning through autonomous situation inference and causal motor-primitive reuse},pdfauthor={Yiyang Jia; Mingyang Lyu; Gongce Cao; Feifei Zhao; Yinqian Sun; Yi Zeng},pdfsubject={Task-agnostic embodied continual learning}}

\definecolor{Positive}{HTML}{3F8F5F}
\definecolor{Accent}{HTML}{0F4D92}
\definecolor{Neutral}{HTML}{68717E}

\newcommand{\model}{SPIRE}
\newcommand{\modelfull}{Situation-guided Primitive Integration for Recurrent Embodiment}

\begin{document}

\title{SPIRE: Task-agnostic embodied continual learning through autonomous
situation inference and causal motor-primitive reuse}

\author{\IEEEauthorblockN{Yiyang Jia\textsuperscript{1,*}, Mingyang Lyu\textsuperscript{2}, Gongce Cao\textsuperscript{3}, Feifei Zhao\textsuperscript{2}, Yinqian Sun\textsuperscript{2}, Yi Zeng\textsuperscript{4}}
\IEEEauthorblockA{\normalfont\small
\textsuperscript{1}University College London, London, United Kingdom\\
\textsuperscript{2}Institute of Automation, Chinese Academy of Sciences (CASIA), Beijing, China\\
\textsuperscript{3}University of Chinese Academy of Sciences (UCAS), Beijing, China\\
\textsuperscript{4}Gaoling School of Artificial Intelligence, Renmin University of China, Beijing, China}
\thanks{*Corresponding author: Yiyang Jia. Email addresses in author order:
\nolinkurl{yiyang.jia.24@ucl.ac.uk},
\nolinkurl{lvmingyang2024@ia.ac.cn},
\nolinkurl{caogongce25@mails.ucas.ac.cn},
\nolinkurl{zhaofeifei0923@163.com},
\nolinkurl{sunyinqian2018@ia.ac.cn},
\nolinkurl{yi.zeng@ruc.edu.cn}.}}
\maketitle
\thispagestyle{plain}

\begin{abstract}
Embodied continual learning requires controllers to infer changing situations,
retain prior behavior and reuse motor skills without task labels or known
boundaries. The challenge is deciding when to recover earlier behavior and
which skills can support new learning. We introduce \modelfull\ (\model),
a recurrent-spiking system that infers latent contexts from action--outcome
evidence and learns context-specific motor actors online. Mature actors become
protected sources that combine with a target-context residual to drive
recurrent-spiking motor populations. Across coordinate,
three-joint and hidden-force control, \model\ maintained control, autonomously
inferred situations, retained prior behavior and accelerated acquisition of an
unseen source composition, reducing mean interactions to criterion by
approximately 40\% relative to matched fresh controllers. Rewinding compatible
source actors impaired early acquisition in every matched intervention,
demonstrating causal reuse. Against seven recurrent continual-learning and
modular-control
comparators, \model\ achieved 2.4 times the mean prior-learning advantage of the
strongest eligible task-free alternative selected for each paired stream.
These results demonstrate autonomous recovery and causal reuse of motor skills
for new learning in continuous embodied control.
\end{abstract}


\section{Introduction}

\begin{figure*}[!t]
  \centering
  \includegraphics[width=0.98\textwidth]{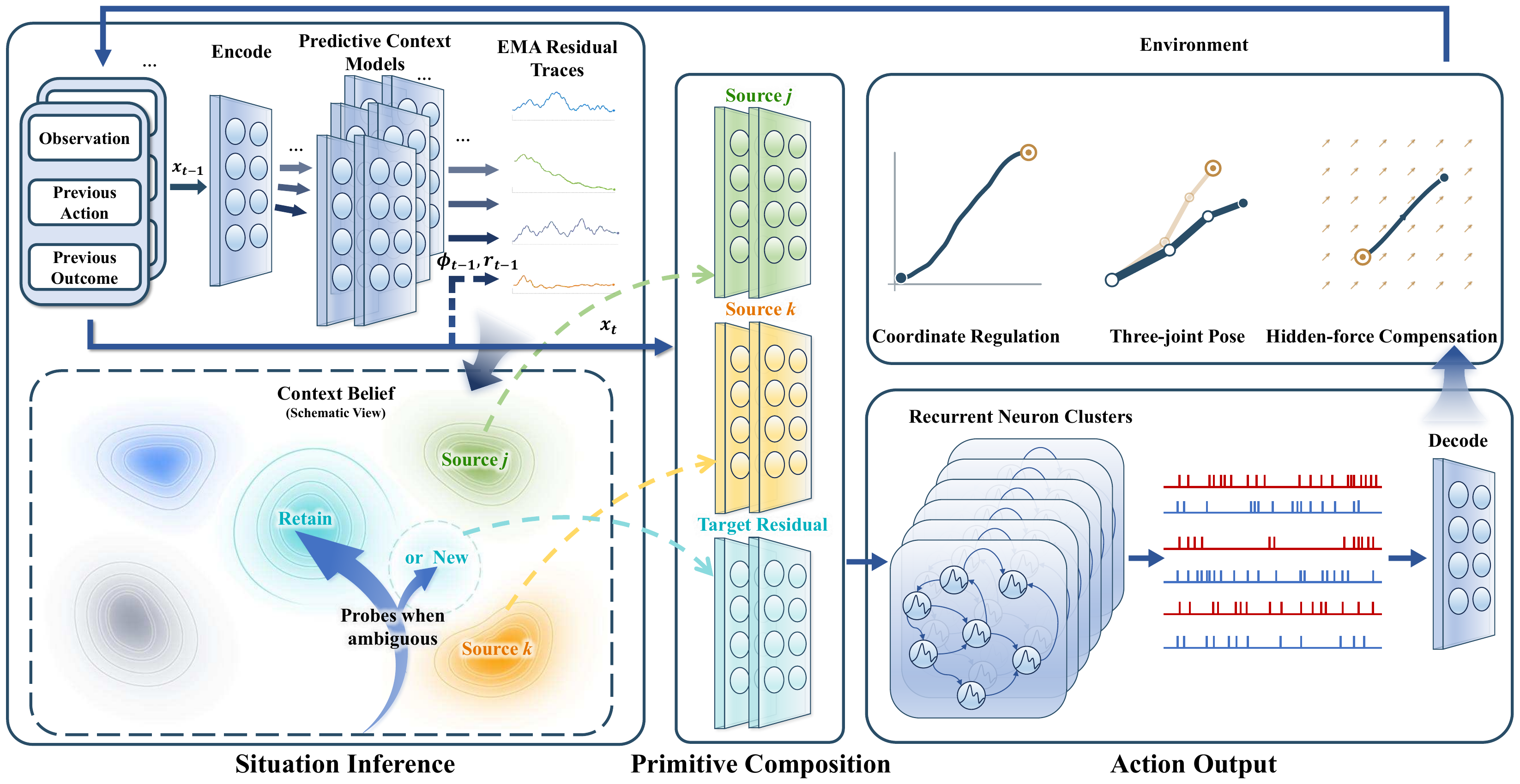}
  \caption{\textbf{SPIRE learning architecture.}
  From left to right: situation inference, primitive composition and
  recurrent-spiking action output. Colored dashed arrows indicate selection
  of source and residual actors, while the solid \(x_t\) arrow carries their
  current input. Paired event rows denote positive and negative population
  channels. Environment feedback supports context inference and local learning.
  The context-belief map and control environments are schematic.}
  \label{fig:v2-overview}
\end{figure*}

Embodied intelligence is learned and expressed inside a closed sensorimotor
loop: actions change the body and environment, their consequences provide
evidence for subsequent adaptation, and learning continues as goals and
dynamics change. Biological motor systems exhibit the corresponding memory and
composition capabilities. In humans, contextual inference regulates which
motor memories are created, updated and expressed, allowing learned behavior to
reappear when its context is inferred again
\citep{heald2021coin,heald2023}. Studies in humans and other animals further
describe movement through flexible combinations of motor primitives and muscle
synergies \citep{thoroughman2000,davella2003,bizzi2008}. This evidence connects
adaptive behavior to context-appropriate memory and reusable action structure,
rather than the isolated acquisition of one control law at a time.

Recurrent reinforcement learning uses interaction history for partially observed
and task-agnostic control \citep{ni2022,caccia2023}. Regularization and replay
reduce interference with prior learning
\citep{kirkpatrick2017,zenke2017,lopezpaz2017,rolnick2019}, but preserving
performance does not determine which skills benefit new learning.
Mixtures of internal models use prediction evidence to select or combine
controllers \citep{wolpert1998mosaic,haruno2001mosaic}. This supports contextual
adaptation and generalization, yet does not establish whether specific retained
skills accelerate new acquisition.

A task-agnostic controller must infer situations from action consequences
without task labels or boundary cues \citep{caccia2023,aljundi2019}.
Since actions shape evidence and identical observations can require different
commands \citep{heald2021coin,wolpert1998mosaic}, it must decide which memories
to express and update while preserving skills for both familiar and new
behaviors. The retained knowledge is therefore closed-loop behavior acquired
through interaction
\citep{caccia2023,bartolozzi2022}.

Inspired by contextual motor memory and reusable motor primitives, we propose
\modelfull\ (\model), comprising three modules: \textbf{situation inference}
recovers familiar behavior from action consequences; \textbf{primitive composition}
reuses protected skills for new learning; and \textbf{action output}
generates recurrent-spiking control commands.
Across coordinate regulation, three-joint pose control and hidden-force
compensation, a seven-baseline comparison shows reliable retention and greater mean
prior-learning advantage than the strongest eligible task-free alternative
selected per stream.

\begin{samepage}
The contributions are threefold:
\begin{enumerate}
  \item \textbf{A framework for task-agnostic continual control}.
  \model\ links three modules to autonomously recover earlier behavior and
  reuse protected skills for new learning without task labels or known boundaries
  (Fig.~\ref{fig:v2-overview}).
  \item \textbf{Quantified transfer across three embodied control forms}.
  \model\ reduces mean interactions to criterion by approximately 40\% relative
  to matched fresh controllers. Its mean prior-learning advantage is 2.4 times
  that of the strongest eligible task-free alternative selected for each paired stream.
  \item \textbf{Mechanistic evidence for selective reuse and spiking control}.
  Source-specific interventions establish selective skill reuse.
  Ablations distinguish the recovery benefits of context organization from
  the acquisition benefits of composition. Motor interventions identify how
  events, recurrence and persistent state contribute to command generation.
\end{enumerate}
\end{samepage}

\section{Related work}

\subsection{Task-agnostic continual learning in embodied control}

Continual-learning methods address interference through EWC/SI-style parameter
regularization, replay and architectural separation
\citep{kirkpatrick2017,zenke2017,rolnick2019,rusu2016}. Task-free variants infer
consolidation events or recruit mixture components from the stream
\citep{aljundi2019,lee2020cndpm}, while contextual gating can isolate neural
populations when context is available \citep{masse2018}. Many benchmarks remain
centered on classification and generation
\citep{parisi2019,delange2022,hadsell2020}. In embodied control, however, actions
make the evidence stream endogenous. Recurrent RL supplies a strong reference
under partial observability \citep{ni2022}, task-agnostic continual RL adapts
without a task-conditioned policy \citep{caccia2023}, and neuromorphic work
embeds adaptive computation in the sensorimotor loop \citep{bartolozzi2022}.
Continued adaptation, however, must be distinguished from recovering a
familiar behavior and using retained skills to accelerate a new one.

\subsection{Latent situation inference from action consequences}

Mixtures of experts assign responsibility among specialized predictors
\citep{jacobs1991}, and MOSAIC uses paired internal models and prediction error
to select or combine motor controllers
\citep{wolpert1998mosaic,haruno2001mosaic}. Latent-cause and change-point models
decide whether evidence supports an existing or new context
\citep{gershman2010,adams2007}, while COIN links contextual inference to the
acquisition and expression of human motor memories
\citep{heald2021coin,heald2023}. These accounts explain how prediction evidence
can guide memory selection. Selecting a familiar controller and identifying
which retained skills benefit an unfamiliar task are nevertheless different
decisions, so contextual responsibility alone does not identify causal reuse.

\subsection{Retention, forward transfer and strict source reuse}

Regularization methods such as EWC and SI reduce movement along important
parameter directions \citep{kirkpatrick2017,zenke2017}. Replay restores earlier
information \citep{lopezpaz2017,rolnick2019}, while modular architectures reserve
task-specific capacity \citep{rusu2016,serra2018}. These mechanisms primarily
protect old solutions, whereas forward transfer asks whether prior knowledge
improves new acquisition. Motor-primitive and muscle-synergy research motivates
reusable action structure \citep{thoroughman2000,davella2003,bizzi2008}, and
lifelong compositional algorithms use shared components to accelerate learning
\citep{ruvolo2013ella,mendez2021}. A learning benefit establishes forward
transfer, but attributing that benefit to particular retained skills requires
source-specific interventions. This distinction is especially important when
both the context and source relevance must be inferred during interaction.

\subsection{Brain-inspired local learning and recurrent-spiking control}

Spiking neural networks provide event-based communication and persistent
temporal state \citep{maass1997,neftci2019}. Three-factor and eligibility-based
rules show how local activity can be combined with a delayed modulatory signal
without exact end-to-end backpropagated gradients
\citep{fremaux2016,bellec2020}. Continual spiking systems use local or
probabilistic adaptation or task-structured population selection
\citep{antonov2022,skatchkovsky2022,han2023,han2025}. These approaches motivate
local adaptation and temporal computation. For embodied control, the
corresponding mechanistic question is how event transmission and recurrent
state affect the commands that act on the environment.

\section{SPIRE learning system}
\label{sec:model}

SPIRE uses action--outcome evidence to determine which behavior to recover,
which source skills to express and where to store new learning.
Figure~\ref{fig:v2-overview} shows the three components that implement these
decisions:
\textbf{situation inference} on the left evaluates action--outcome evidence to maintain,
retrieve or recruit a latent context; \textbf{primitive composition} in the middle uses
that context to combine protected source actors with a target-context residual
actor; and \textbf{action output} on the right converts the composed motor drive into
recurrent population events and a control command. The section follows these
three components and then describes learning from the returned feedback.
Shared controller constants are listed in \ref{app:v2-controller}.

One control step, or transaction, consists of receiving an observation,
executing an action and receiving its resulting observation and scalar outcome.
The figure distinguishes two information paths: feedback from the completed
step \(t-1\) supplies evidence for context inference, while the current input
\(x_t\) directly supplies the actors in the middle. Context selection determines
which actors are expressed; their current input determines the motor drive.
Feedback from action \(a_t\) becomes evidence for subsequent control and updates
the context that generated that action.

\subsection{Situation inference}
\label{sec:v2-situation}

The left component follows Encode, Predictive Context Models,
EMA Residual Traces and Context Belief. Prediction evidence and,
when needed, motor probes determine the active latent-context slot, an index
into persistent context-specific memory.

\textbf{Situation evidence combines observation, action and outcome.}
At step \(t\), the controller receives the normalized observation
\(o_t\in[-1,1]^{d_o}\), its internally retained previous action
\(a_{t-1}\in[-1,1]^{d_a}\), and the previous scalar outcome
\(r_{t-1}\in[-1,1]\):
\begin{equation}
  x_t = [o_t,a_{t-1},r_{t-1}].
  \label{eq:v2-stream-input}
\end{equation}
The same input supplies both the encoder and the actors. A fixed encoder
preserves leading input coordinates and adds projected coordinates where
needed to form a 12-dimensional representation \(z_t\), then constructs a normalized
16-dimensional activity basis \(h_t\). This basis supplies the predictive
models; context belief instead compares evidence for the six available
latent-context slots. The encoding equations are given in
\ref{app:v2-representations}.

\textbf{Protected context models predict transitions and outcomes.}
Each mature latent-context slot \(i\) with a valid motor--outcome signature
stores a protected utility template \(\bar w_i\) and action-conditioned forward
template \(\bar F_{i,k}\), captured from its learned utility and forward models.
To evaluate the completed step \(t-1\), SPIRE constructs a transition-feature
vector \(\phi_{t-1}\) from the observed state, its change and the executed
action, including squared terms. Its \(d_c=1+4d_o+2d_a\) entries are clipped
to \([-1,1]\); the full expansion appears in
\ref{app:v2-representations}. The saved encoding \(h_{t-1}\) and
motor-drive index \(k_{t-1}\) identify the input and action condition for each
template:
\begin{equation}
  \widehat r_{t-1}^{(i)}=\bar w_i^{\mathsf T}\phi_{t-1},\qquad
  \widehat\phi_{t-1}^{(i)}=h_{t-1}^{\mathsf T}\bar F_{i,k_{t-1}}.
  \label{eq:v2-context-predictions}
\end{equation}
Here \(k_{t-1}\) is the nearest entry in a fixed signed motor-drive codebook,
recorded when the action was generated. The utility template estimates the
outcome associated with the realized transition, while the forward template
estimates that transition from the action-time encoding. Comparing both with
the returned evidence gives the absolute residual vector
\(e_{t-1}^{(i)}=[|r_{t-1}-\widehat r_{t-1}^{(i)}|,
|\phi_{t-1}-\widehat\phi_{t-1}^{(i)}|]\).

\textbf{Accumulated prediction errors define context belief.}
For each calibrated context, an exponential moving average (EMA) summarizes
the successive residuals:
\begin{equation}
  E_{t-1}^{(i)}=0.90E_{t-2}^{(i)}+0.10e_{t-1}^{(i)}.
  \label{eq:v2-residual-ema}
\end{equation}
The first available residual initializes the trace. Each context calibrates
its ordinary residual mean \(\mu_i^{\mathrm{err}}\) and scale
\(\sigma_i^{\mathrm{err}}\) from 64 local samples, with a scale floor of 0.02.
The mean positive standardized residual defines its score and belief:
\begin{equation}
  \begin{aligned}
  S_{i,t-1}&=\operatorname{mean}\!\left[
    \max\!\left(\frac{E_{t-1}^{(i)}-\mu_i^{\mathrm{err}}}
      {\sigma_i^{\mathrm{err}}},0\right)\right],
  \\
  b_{i,t-1}&=\frac{\exp(-S_{i,t-1})}
    {\sum_{j\in\mathcal C}\exp(-S_{j,t-1})},
  \end{aligned}
  \label{eq:v2-context-belief}
\end{equation}
where division and the maximum are element-wise and \(\mathcal C\) contains
the calibrated contexts. A smaller score indicates a closer fit to the
context's usual prediction error; belief expresses relative support for
familiar contexts.

\textbf{Motor probes confirm potential context changes.}
If the active context continues to explain the stream, it remains selected.
A persistently better familiar alternative, sustained unexplained residual or
context-relative outcome deficit requests signed motor probes for a subsequent
control step. Belief updates alone do not switch the active actor.

SPIRE also runs change-detection probes when a stable context encounters the
aliased reset observation. At this point distinct hidden regimes can have the
same observation, so observation history alone contains no discriminating
sample. On successive control steps, SPIRE sets one component of a probe motor
drive \(u_n^{\mathrm{probe}}\) to \(+0.45\) or \(-0.45\), alternates the sign
and action axis, and sends that drive through the same recurrent-spiking motor
used for ordinary control. It accumulates
\begin{equation}
  s_d=\frac{1}{N}\sum_{n=1}^{N}
  r_n\operatorname{sign}(u_{n,d}^{\mathrm{probe}}),
  \qquad N\leq24.
  \label{eq:v2-signature}
\end{equation}
The vector \(s\in\mathbb{R}^{d_a}\) records how the observed outcome changes
under positive and negative intervention along each action dimension. While
these change-detection probes run, the probe drive replaces the ordinary
policy drive and actor, critic, utility and forward-model learning is paused.

\textbf{Signature matching selects a familiar or new context.}
At the aliased reset observation, acquisition and retrieval are state matched,
so SPIRE uses the strict stored-signature distance. After at least two balanced
positive--negative pairs per action axis, a confident familiar match ends the
probes early. Otherwise SPIRE continues for at most 24 probe steps. Signature
matching retains the current slot, retrieves another familiar slot or, if no
match exists and capacity remains, recruits a new slot. Probes triggered by
predictive history away from the aliased reset observation may use a bounded
state-tolerant match, while a credible novel source composition takes precedence
over state-tolerant familiar-context retrieval.

The selected slot determines the actor and learning state for subsequent
steps. The Retain and New regions in the figure schematically
represent familiar-context recovery and recruitment based on these residuals
and signatures. This use of action consequences follows internal-model and
contextual-inference accounts of motor learning
\citep{wolpert1995,haruno2001mosaic,heald2021coin}.

\subsection{Primitive composition and reuse}
\label{sec:v2-composition}

New-context learning draws on compatible protected skills while updating a
separate target actor. The middle component receives the selected context and
the current input along the solid \(x_t\) arrow. Its three actor stacks, Source \(j\),
Source \(k\) and Target Residual, produce a combined motor drive
for the recurrent neuron clusters on the right.

For active latent-context slot \(i\), the actor receives the current input
with a bias term, \(f_t=[1,x_t]\), and proposes
\begin{equation}
  \mu_{i,t}=\tanh(A_i f_t).
  \label{eq:v2-actor}
\end{equation}
Once a context-specific actor is mature, its weights are copied to a protected
source actor \(A_i^{\mathrm{src}}\), a read-only copy for other contexts. These
actors provide the reusable motor primitives. The active actor \(A_i\) learns
the target residual when sources are expressed; without an active source pair,
its proposal \(\mu_{i,t}\) supplies the deterministic motor drive directly.

When a new latent-context slot is recruited, every pair of mature protected
source actors \((j,k)\) is scored by the alignment between its summed source
signature and the new signature:
\begin{equation}
  q(j,k\mid s)=
  \frac{(s_j+s_k)^{\mathsf T}s}
  {\lVert s_j+s_k\rVert\lVert s\rVert}.
  \label{eq:v2-source-score}
\end{equation}
The best pair is selected when \(q\geq0.75\) and, if another pair exists,
exceeds the runner-up by at least 0.20. This signature comparison determines
which source contexts are linked to the middle component by the colored
dashed arrows. Their contributions are allocated separately along each action
dimension:
\begin{equation}
  m_{j,d}=\frac{|s_{j,d}|}{|s_{j,d}|+|s_{k,d}|},
  \qquad m_{k,d}=1-m_{j,d}.
  \label{eq:v2-source-mask}
\end{equation}
If both source magnitudes are zero, the two masks are equal. The selected pair
and masks are stored with the target slot. The deterministic motor drive becomes
\begin{equation}
  \bar u_{i,t}=\operatorname{clip}\!\left(
  \sum_{p\in\{j,k\}}m_p\odot
  \tanh(A_p^{\mathrm{src}}f_t)
  +0.3\mu_{i,t},-1,1\right).
  \label{eq:v2-composition}
\end{equation}
The two protected source actors provide dimension-wise source structure while
the active actor contributes the target-context residual. Updates made while
learning the target context do not modify either protected source actor. The
composition expresses the functional role of reusable motor primitives
\citep{thoroughman2000,davella2003,bizzi2008}.

During ordinary transactions, SPIRE samples a bounded signed perturbation
\(\epsilon_t\in\{-\sigma_t,+\sigma_t\}^{d_a}\) and sends
\begin{equation}
  u_t^*=\operatorname{clip}(\bar u_{i,t}+\epsilon_t,-1,1)
  \label{eq:v2-policy-drive}
\end{equation}
to the recurrent-spiking motor along the arrow from the primitive stacks to
the neuron clusters. The perturbation supports online actor learning
(Section~\ref{sec:v2-local-learning}).

\begin{figure*}[!t]
  \centering
  \includegraphics[width=0.98\textwidth]{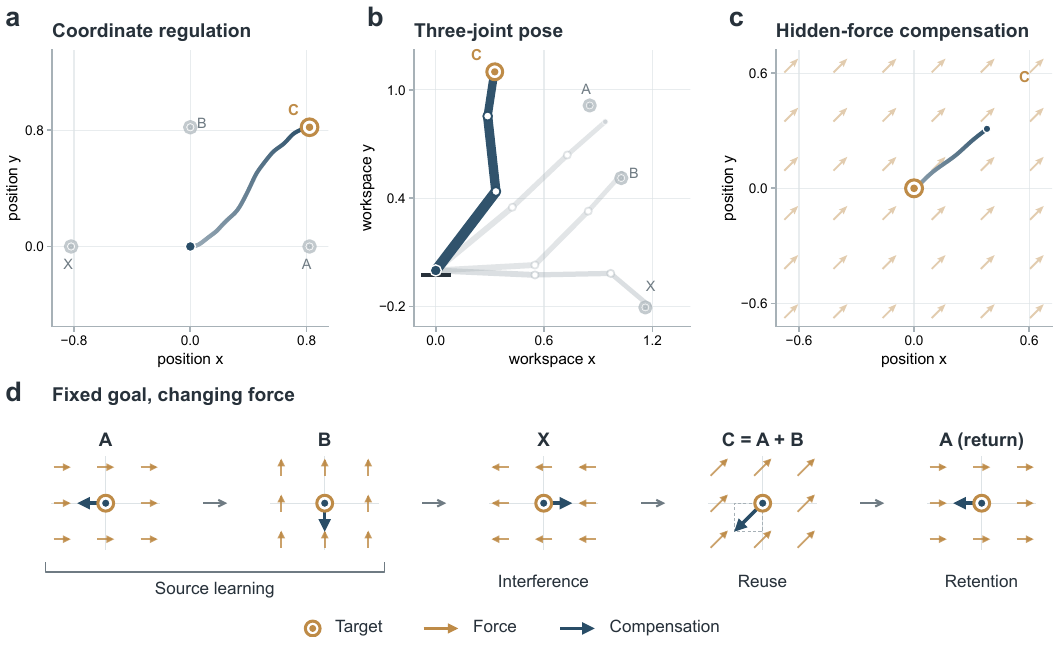}
  \caption{\textbf{Three embodied control forms and the continuous task sequence.}
  \textbf{(a)} Coordinate regulation uses axial source targets A and B, an
  opposing target X and the unseen diagonal composition C.
  \textbf{(b)} Joint control uses one three-link body and base. Gray poses
  illustrate A, B and X, and the navy pose illustrates C; rings mark ideal
  target endpoints, with each arm shown near its target.
  \textbf{(c)} Hidden-force compensation stabilizes the body at the origin
  against the spatially uniform C disturbance \((+x,+y)\).
  \textbf{(d)} The sequence is illustrated with hidden forces while the origin
  target remains fixed: source skills are learned in A and B, X introduces
  interference, C combines the A and B disturbances to test reuse, and
  return-A tests retention. Target rings, force arrows and compensation
  arrows follow the shared legend. Paths and compensation directions are
  schematic. Controller memory persists throughout the sequence, and phase
  labels are not supplied to SPIRE.}
  \label{fig:v2-tasks}
\end{figure*}

\subsection{Action output}
\label{sec:v2-spike}

The right component follows Recurrent Neuron Clusters, paired population
events, Decode and Environment. It receives \(u_t^*\) from the
primitive composition and outputs the command \(a_t\).

The motor drive is converted to binary events over 12 neural substeps. Each of
the \(d_e\) motor-drive dimensions contains a population of \(K=7\) neurons in a
fixed symmetric bidirectional ring. For dimension \(j\), neuron \(k\) and
neural substep \(n\), let \(\widetilde c_{j,k}^{(n)}\) be the mean recurrent
state of its two immediate neighbors, with periodic indexing. Membrane,
signed event and recurrent state follow
\begin{equation}
\begin{aligned}
  V_{j,k}^{(n+1)} &=
  0.70V_{j,k}^{(n)}+u_{t,j}^*+0.03\widetilde c_{j,k}^{(n)},\\
  e_{j,k}^{(n)} &= \mathbb{1}[V_{j,k}^{(n+1)}\geq0.78]
            -\mathbb{1}[V_{j,k}^{(n+1)}\leq-0.78],\\
  c_{j,k}^{(n+1)} &=0.45c_{j,k}^{(n)}+e_{j,k}^{(n)}.
\end{aligned}
  \label{eq:v2-motor-dynamics}
\end{equation}
After an event, the corresponding signed threshold is subtracted from the
membrane. All membrane and recurrent states persist across control steps.

The implementation stores two binary arrays,
\(e^{+(n)},e^{-(n)}\in\{0,1\}^{d_e\times K}\), and forms
\(e^{(n)}=e^{+(n)}-e^{-(n)}\). Positive and negative events are carried by
opponent population-event channels: one positive--negative pair for each
motor-drive dimension, represented by the paired event rows in the figure.
The fixed decoder pools events from every neuron and all 12 substeps:
\begin{equation}
  a_t=D\,\operatorname{clip}\!\left(
  \frac{1}{12K}\sum_{n=1}^{12}\sum_{k=1}^{K}e_{\cdot,k}^{(n)},-1,1\right).
  \label{eq:v2-decoder}
\end{equation}
For all three control forms, \(d_e=d_a\) and \(D=I\). The population event
count is the decoder's complete input: suppressing both event arrays produces
a zero control command. The environment executes \(a_t\) and returns
\((o_{t+1},r_t)\), from which SPIRE constructs \(\phi_t\) by the same rule as
Eq.~\eqref{eq:v2-consequence}. This return path supplies the left component with
new inference evidence and the local updates described next. Across the three
control forms, observation, action, motor-drive and transition-feature widths
remain fixed within each stream; body dimensions change the tensor shapes
while the equations and learning coefficients remain the same.

\subsection{Context-local learning}
\label{sec:v2-local-learning}

Environment feedback trains the context-specific models that support inference
on the left and the actor expressed in the middle. Each latent-context slot
\(i\) indexes its actor \(A_i\), critic \(v_i\), utility vector \(w_i\),
forward model \(F_i\) and motor--outcome signature \(s_i\).

When generating an action, SPIRE temporarily records its originating context,
\(f_t\), \(h_t\), \(k_t\) and sampled perturbation \(\epsilon_t\) in an action
token. After \((o_{t+1},r_t)\) returns, the saved \(h_t,k_t\) are used to
evaluate the protected predictive templates. Local learning then uses the
context and perturbation in that same token, and the residual evidence updates
the belief for subsequent control. This assignment keeps an action's learning
attached to the context that generated it even when the new evidence requests
context confirmation.

The critic estimates the outcome expected from the current input. Its error,
\(\delta^V_t=r_t-v_i^{\mathsf T}f_t\), modulates the actor update:
\begin{equation}
  \Delta A_i =
  \eta_A\,
  \frac{\delta^V_t\,\epsilon_t f_t^{\mathsf T}}
  {\Gamma_{i,t}(0.1+\lVert f_t\rVert^2)},
  \qquad \eta_A=0.08.
  \label{eq:v2-actor-update}
\end{equation}
The sampled perturbation \(\epsilon_t\) links the observed outcome to the
action, while \(f_t\) identifies the actor inputs responsible for that action.
The normalization \(\Gamma_{i,t}\) accounts for exploration amplitude.
Exploration decreases after the context assignment stabilizes, with the
normalization preserving the acquisition-scale update gain. The exact scaling
and critic update are given in \ref{app:v2-auxiliary-learning}.

The utility model learns which scalar outcome accompanies the realized
transition features \(\phi_t\). The forward model indexed by the saved drive
code \(k_t\) learns to predict those features from \(h_t\). Both use their
prediction errors in normalized local updates
(\ref{app:v2-auxiliary-learning}). All actor, critic, utility and
forward updates apply to the slot stored in the action token; other slots
retain their state. The protected predictive templates retain their captured
weights while the context-specific models continue learning.

A latent-context slot becomes mature after 64 local updates. At that point its
current context-specific actor is copied to the protected source actor
\(A_i^{\mathrm{src}}\). Later updates by the same slot refresh this read-only
source copy, while other slots can only read it. Maturity makes the actor
available for reuse. Context stability additionally requires sustained
prediction accuracy and positive outcomes; it governs probe eligibility and
the reduced exploration scale. Its exact conditions are given in
\ref{app:v2-auxiliary-learning}.

\section{Experimental design}
\label{sec:experiments}

\subsection{Experimental questions}

We evaluate SPIRE through three questions:
(1) \textbf{Control and retention}: can it maintain control and recover earlier
behavior after interference without task labels or known boundaries?
(2) \textbf{Learning benefit}: does prior experience accelerate acquisition of
a new task, and how does this benefit compare with task-free alternatives?
(3) \textbf{Mechanistic contributions}: which mechanisms support recovery and
new-task learning, and which retained skills cause the learning benefit?

\subsection{Three embodied control forms}

We evaluate the same model in three simulated control forms, implemented
for a shared source-composition protocol and motivated by internal-model
motor control and force-field adaptation
\citep{wolpert1998mosaic,shadmehr1994}.
(1) \textbf{Coordinate regulation} observes planar position and velocity and
outputs two acceleration commands (Fig.~\ref{fig:v2-tasks}a).
Tasks A, B and X target \((+x)\), \((+y)\) and \((-x)\), while C targets \((+x+y)\).
(2) \textbf{Three-joint pose control} observes three joint positions, three
velocities and a planar endpoint, and emits three joint commands
(Fig.~\ref{fig:v2-tasks}b). A and B control the first two joints,
X controls the third in the opposite direction, and C combines the first two.
(3) \textbf{Hidden-force compensation} observes planar position and velocity
and emits two compensatory commands (Fig.~\ref{fig:v2-tasks}c).
Its hidden disturbances follow the same signed
composition, while the behavioral objective remains stabilization at the
origin.

Each form uses independent controller instances with the same learning rules
and non-dimensional coefficients. All observation, action and transition-feature
channels are normalized. The scalar outcome combines normalized progress toward
the current objective, control quality and an action-energy cost. Interface
dimensions are in \ref{app:v2-controller}, Table~\ref{tab:v2-forms}.
System equations and function thresholds are in \ref{app:v2-thresholds}.

\subsection{Continuous protocol and matched comparisons}

Within each form, we test continual learning using one uninterrupted
\(A\rightarrow B\rightarrow X\rightarrow C\rightarrow A_{\mathrm{return}}\)
stream (Fig.~\ref{fig:v2-tasks}d). Each phase contains five 128-transaction
trials. At a trial boundary,
the evaluator returns the body to its ordinary resting state, but controller
parameters, latent-context memory, previous action and outcome, and recurrent
private state remain continuous. Phase labels are used only to configure the
environment and score behavior; they are never supplied to SPIRE. The first sample
after a situation change is the shared aliased reset observation, so context
recovery must use retained sensorimotor history and, when needed, bounded signed
motor probes.

Retention compares initial and return-A performance on the uninterrupted
stream, while context recovery tests retrieval of the original A context.
For new learning, the experienced controller enters C after A, B and X.
A fresh controller with the same initialization starts C from the same
environment state, with random states aligned at entry.

Source-specific interventions test which retained skills cause the learning
benefit. Once SPIRE recruits a C context and selects its source pair, the
controller, environment and random state are copied into six matched
continuations: intact, exact sham, and rewinding A+B, A alone, B alone or X.
Only the selected protected source actors are removed; the target residual,
utility and forward models, and remaining experience are preserved. A loss
after removing A or B but little change after removing X identifies selective
reuse of compatible skills.

\begin{figure*}[!tp]
  \centering
  \includegraphics[width=0.78\textwidth]{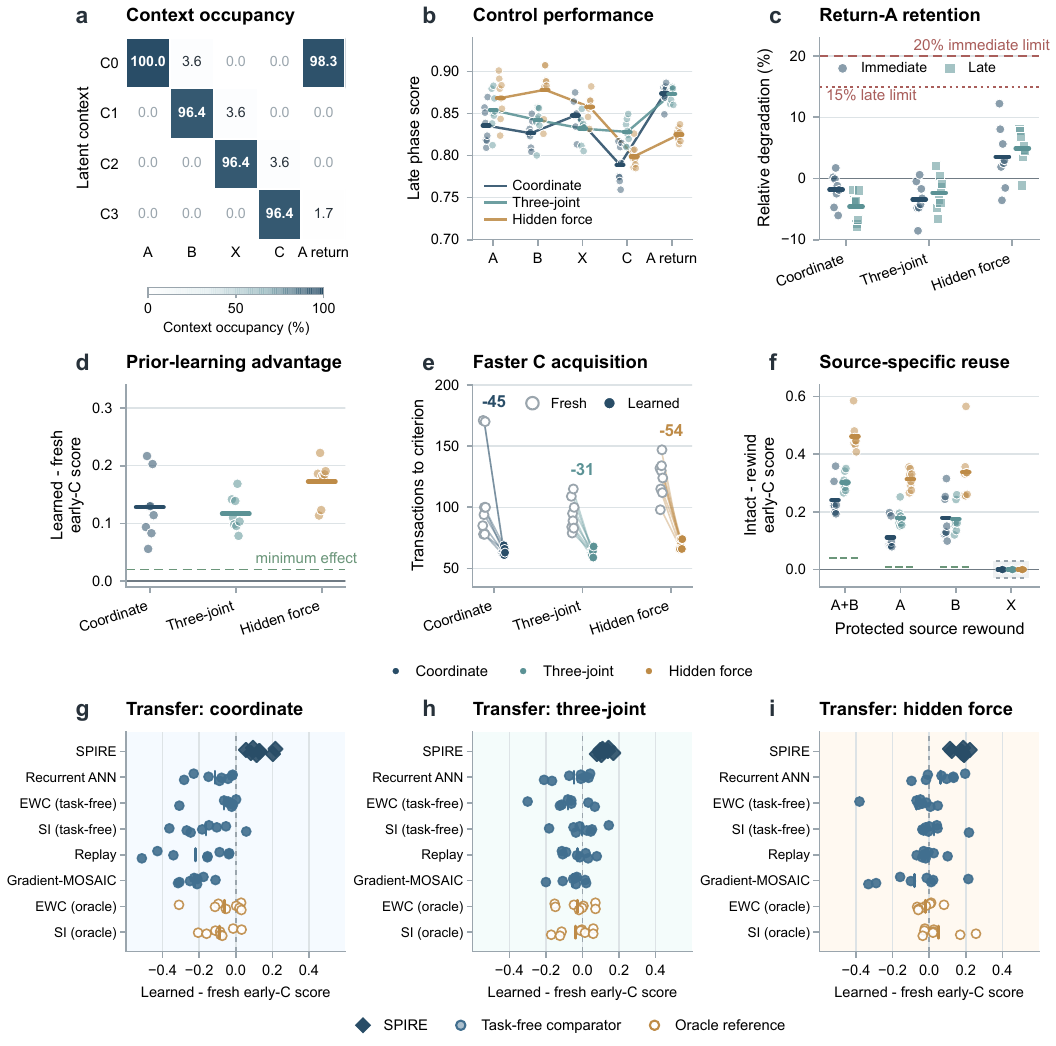}
  \caption{\textbf{Context inference, control, retention and source reuse.}
  Each form has eight replicates per method; points show individual runs and
  short bars show means.
  \textbf{(a)} Mean occupancy of every observed latent context across 24 SPIRE
  runs. Each phase includes all 640 assignments per run, and each column sums
  to 100\%; return-A recovers C0.
  \textbf{(b)} Late phase scores across A--B--X--C--A, with form means connected.
  \textbf{(c)} Immediate and late return-A degradation; negative values indicate
  improvement and dashed lines mark the retention limits.
  \textbf{(d)} Early-C score advantage over a same-initialization fresh controller.
  \textbf{(e)} Transactions to the C criterion; lines pair learned and fresh
  runs, with mean reductions annotated.
  \textbf{(f)} Early-C score loss after rewinding A+B, A, B or X source actors.
  Green marks show reuse thresholds; shading marks the X tolerance of
  \(\pm0.03\).
  \textbf{(g--i)} Prior-learning advantages across methods in coordinate, joint
  and force control. Oracle EWC and SI receive phase-start information.}
  \label{fig:v2-results-overview}
\end{figure*}

\subsection{Comparators and paired execution}

We compare SPIRE with seven baselines covering recurrent adaptation,
parameter consolidation, replay and modular control:
(1) \textbf{naive recurrent ANN}, an online actor--critic testing adaptation
through recurrent memory \citep{ni2022};
(2) \textbf{task-free EWC} and (3) \textbf{task-free SI}, which test parameter
protection \citep{kirkpatrick2017,zenke2017} with outcome-triggered consolidation;
(4) \textbf{Replay}, a bounded reservoir of past policy gradients that tests
rehearsal \citep{rolnick2019};
(5) \textbf{Gradient-MOSAIC}, a four-module implementation of the published
forward--inverse model equations \citep{wolpert1998mosaic,haruno2001mosaic}
that tests prediction-guided modular control;
(6) \textbf{oracle EWC} and (7) \textbf{oracle SI}, which use the same penalties
but receive true phase starts as privileged retention references.
The recurrent ANN variants share a 128-unit online actor--critic. Baseline
settings and implementation-specific adaptations are given in
\ref{app:v2-baselines}.

All methods receive the same normalized observation and scalar outcome.
The first five baselines form the eligible task-free comparison; oracle EWC
and SI are reported separately because of their phase-start information.
Environment seeds are paired across methods within each form and replicate,
while controller seeds are unique across method rows. The evaluation contains
three forms, eight methods and eight paired replicates
(\(3\times8\times8=192\) runs), comprising 825,840 transactions.
Replication and information-access details are in
\ref{app:v2-replication}.

\subsection{Evaluation measures}
\label{sec:v2-measures}

We assess four aspects: control and context recovery, retention, new-task
learning, and source/motor intervention effects.

\textbf{Control and context recovery.} Mean score and success fraction over the
final 256 transactions of each phase, using the form-specific thresholds in
\ref{app:v2-thresholds}, measure control performance. In
Fig.~\ref{fig:v2-results-overview}a, each column shows mean context occupancy
over the 640 assignments per run in that phase, including assignments during
transitions. Comparing the dominant rows in the initial-A and return-A
columns tests whether the same context was recovered; comparing A, B, X
and C tests whether the controller distinguished the four situations.

\textbf{Retention.} Percentage degradation is
\(100(S_A-S_{\mathrm{return}})/\max(|S_A|,10^{-9})\), where \(S_A\) is the
mean score over the final 256 transactions of initial A. Return scores use
the first 128 and final 256 transactions of return-A, with respective limits
of 20\% and 15\%. Lower values are better, and negative values indicate improvement.

\textbf{New-task learning.} The prior-learning advantage is the learned-minus-fresh mean score over the
first 192 C transactions, with a minimum effect of 0.02. Acquisition time
is the first transaction at which the 32-transaction moving-mean score
reaches 0.70, 0.68 or 0.73 for coordinate, joint or force control,
respectively. The same threshold applies to learned and fresh controllers;
larger score advantages and shorter acquisition times indicate better learning.

\textbf{Intervention effects.} Source-specific reuse is
measured by the intact-minus-rewind mean score over the first 192 available
transactions of the matched C continuations. The effect thresholds are 0.04
for jointly rewinding A and B and 0.01 for rewinding either source alone,
with an absolute tolerance of 0.03 for rewinding X. The sham continuation
provides an unchanged reference. Motor dependence is measured by the mean
absolute change in decoded commands under matched output-event suppression,
recurrence removal or private-state reset, using the 32-step procedure in
\ref{app:v2-motor-interventions}.

\section{Results}
\label{sec:results}

Across all 24 streams, SPIRE maintained control, inferred changing situations,
retained earlier behavior and reused protected source actors to learn the
unseen composition. Figure~\ref{fig:v2-results-overview} brings these results
together with the comparison against seven continual-learning alternatives.

\subsection{Autonomous situation inference with sustained control}

\begin{figure*}[!t]
  \centering
  \includegraphics[width=0.97\textwidth]{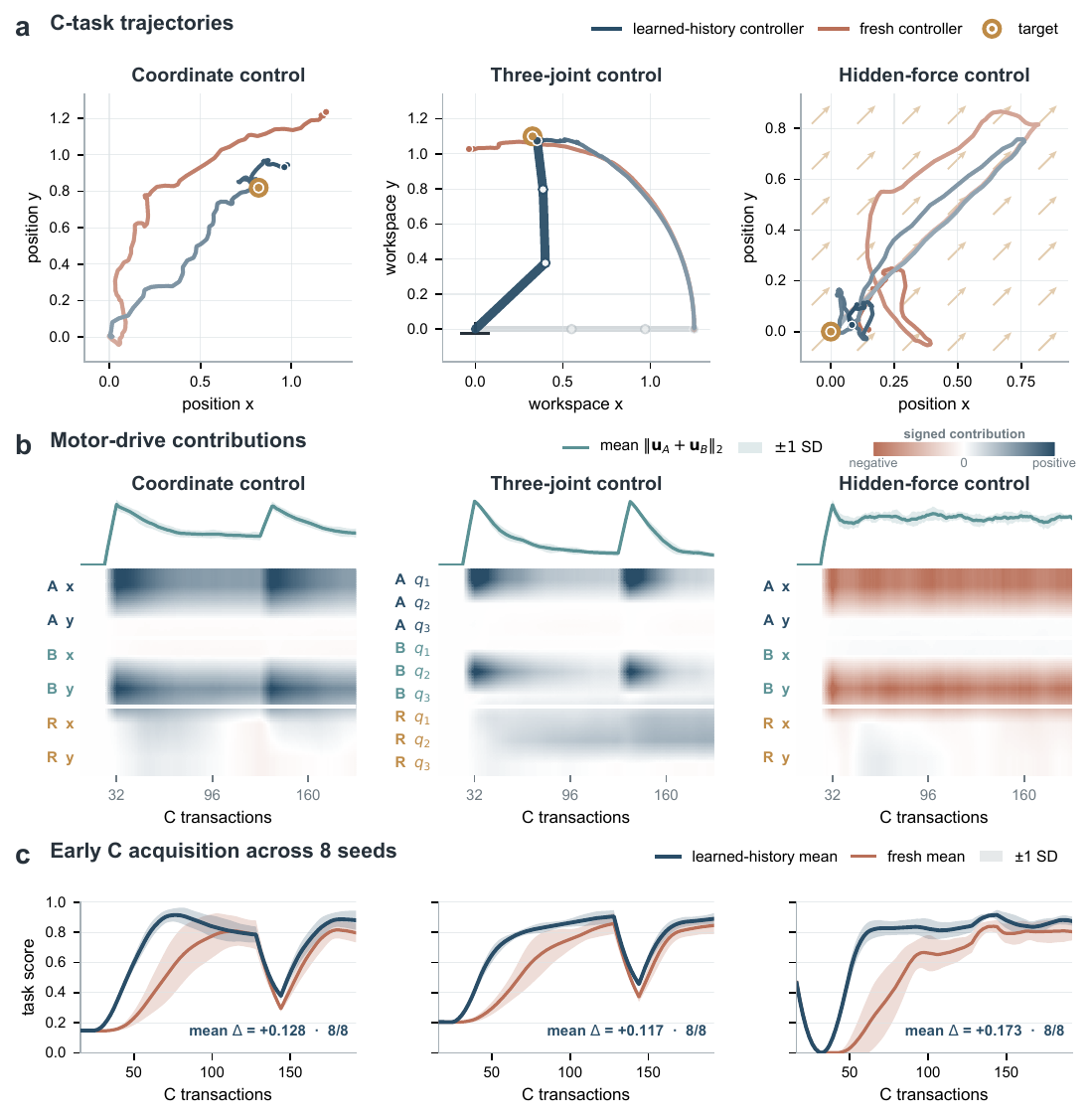}
  \caption{\textbf{Protected source contributions during C acquisition.}
  Each form contains eight matched learned/fresh pairs.
  \textbf{(a)} Selected high-benefit examples over the first 128 C transactions:
  learned-history controller (blue) and same-initialization fresh controller
  (coral). Saturation indicates elapsed time; dots mark endpoints. Gray and
  blue arm poses show the initial and learned-history final three-joint pose.
  \textbf{(b)} Protected source A/B and active-context actor R contributions over
  192 C transactions, before clipping, exploration and recurrent event
  decoding; R becomes the target residual once composition is selected.
  Signed confirmation probes contribute no actor activity.
  Eight-transaction moving means are averaged across replicates. Each form
  uses a symmetric color scale at the 98th percentile of absolute mean
  contributions: blue is positive, coral negative and white zero. The upper
  trace shows mean \(\lVert\mathbf{u}_A+\mathbf{u}_B\rVert_2\) with a
  \(\pm1\) sample-SD envelope.
  \textbf{(c)} Learned/fresh scores: 16-transaction moving means with
  \(\pm1\) sample-SD envelopes, clipped to the score range [0,1]. Annotations
  give unsmoothed mean paired gaps and positive-pair counts.}
  \label{fig:v2-reuse-process}
\end{figure*}

Across the uninterrupted A--B--X--C--A stream, latent-context assignments
changed autonomously. A, B, X and C occupied four distinct modal contexts, and the final A
phase recovered the context used for the initial A phase
(Fig.~\ref{fig:v2-results-overview}a). Because no task, phase or boundary signal entered
the controller, this sequence arose from sensorimotor history rather than an
externally supplied task index. Initial A occupied context 0 throughout. B,
X and C spent 96.4\% of their transactions in contexts 1, 2 and 3,
respectively; return-A recovered context 0 for 98.3\% of its transactions.

During these changes, SPIRE passed the form-specific function criterion in
every phase and replicate. Late scores remained high in coordinate, joint
and hidden-force control as objectives and dynamics changed
(Fig.~\ref{fig:v2-results-overview}b).

\begin{table*}[!tp]
  \centering
  \caption{\textbf{Return-A degradation across methods (\%).}
  Mean \(\pm\) sample SD of score loss from initial-A final 256 transactions
  to return-A first 128 (immediate) or final 256 (late), across eight runs per
  method and form. Negative values indicate improvement.
  Bold values mark the lowest
  mean among task-free methods in each column.
  The two oracle references above the rule receive phase-start information.}
  \label{tab:v2-retention-transfer}
  \small
  \setlength{\tabcolsep}{0.5pt}
  \renewcommand{\arraystretch}{1.12}
  \begin{tabular*}{\linewidth}{@{\extracolsep{\fill}}lcccccc@{}}
\toprule
Method & \multicolumn{2}{c}{Coordinate} & \multicolumn{2}{c}{Three-joint} & \multicolumn{2}{c}{Hidden force} \\
\cmidrule(lr){2-3}\cmidrule(lr){4-5}\cmidrule(lr){6-7}
 & Immediate & Late & Immediate & Late & Immediate & Late \\
\midrule
Oracle EWC & \(13.1 \pm 7.1\) & \(-5.2 \pm 2.2\) & \(10.4 \pm 5.6\) & \(-4.7 \pm 2.4\) & \(2.4 \pm 2.7\) & \(-5.3 \pm 3.1\) \\
Oracle SI & \(9.3 \pm 3.4\) & \(-6.1 \pm 1.6\) & \(12.7 \pm 7.4\) & \(-4.2 \pm 1.6\) & \(3.8 \pm 4.4\) & \(-5.2 \pm 2.4\) \\
\cmidrule(lr){1-7}
\shortstack[l]{Naive recurrent\\ANN} & \(8.5 \pm 7.1\) & \(-6.5 \pm 2.0\) & \(9.2 \pm 3.6\) & \(-4.0 \pm 1.8\) & \(6.4 \pm 4.7\) & {\boldmath \(-4.7 \pm 1.5\)} \\
Task-free EWC & \(9.2 \pm 4.5\) & \(-1.0 \pm 4.1\) & \(14.0 \pm 7.2\) & \(-3.0 \pm 2.0\) & \(3.1 \pm 2.2\) & \(-4.2 \pm 1.5\) \\
Task-free SI & \(5.7 \pm 7.1\) & \(-7.4 \pm 2.9\) & \(7.4 \pm 3.4\) & \(-4.9 \pm 1.4\) & \(5.5 \pm 3.6\) & \(-4.4 \pm 1.7\) \\
Replay & \(9.4 \pm 4.3\) & \(-6.2 \pm 2.9\) & \(13.5 \pm 6.8\) & \(-2.6 \pm 1.1\) & {\boldmath \(1.4 \pm 2.3\)} & \(-4.4 \pm 1.5\) \\
\shortstack[l]{Gradient-\\MOSAIC} & \(16.8 \pm 5.1\) & {\boldmath \(-11.1 \pm 4.0\)} & \(17.7 \pm 5.5\) & {\boldmath \(-6.1 \pm 2.7\)} & \(15.2 \pm 3.7\) & \(-3.1 \pm 1.6\) \\
\shortstack[l]{\textbf{SPIRE}\\\textbf{(ours)}} & {\boldmath \(-1.8 \pm 2.6\)} & \(-4.6 \pm 2.4\) & {\boldmath \(-3.4 \pm 2.9\)} & \(-2.3 \pm 2.9\) & \(3.5 \pm 5.1\) & \(4.9 \pm 2.9\) \\
\bottomrule
\end{tabular*}

\end{table*}

\subsection{Earlier behavior remained recoverable after interference}

To test whether learning B, X and C disrupted an earlier skill, we compared
return-A performance with the final learning window of initial A. The
immediate return window tests recovery before extensive re-exposure, while
the late window tests whether performance remains stable thereafter.
All 24 SPIRE runs met both retention criteria
(Fig.~\ref{fig:v2-results-overview}c). The largest late drop was 8.08\%, below
the 15\% limit. Together with recovery of the original A context, these
results show that the earlier behavior remained accessible after learning
the new composition, rather than being replaced by subsequent experience.

\subsection{Prior learning accelerated acquisition through source-specific reuse}

The learned-versus-fresh C comparison was positive in all 24 SPIRE runs
(Fig.~\ref{fig:v2-results-overview}d). Mean early-C advantages were 0.128 in
coordinate control, 0.117 in joint control and 0.173 in force control, with a
minimum of 0.056 across replicates. Earlier experience
therefore improved acquisition of C across all three forms under matched
initialization and aligned C-entry randomness.

The same benefit appears in acquisition time. Relative to fresh controllers,
controllers with A--B--X history reached the C criterion 44.8 transactions
earlier on average in coordinate control, 30.9 earlier in joint control and
54.1 earlier in force control (Fig.~\ref{fig:v2-results-overview}e).
Pooled mean interactions to criterion fell from 109.25 for matched fresh
controllers to 66.00 for experienced controllers, a 39.6\% reduction.
The learned and fresh score curves in Fig.~\ref{fig:v2-reuse-process}c
show how this advantage develops during acquisition.

\textbf{Source-specific interventions identify the cause of transfer.}
A positive prior-learning advantage alone does not identify which prior knowledge produced the
benefit. Selectively rewinding the protected A and B source actors reduced early-C
performance in every run, with mean intact-minus-rewind effects of 0.240,
0.302 and 0.462 for coordinate, joint and force control
(Fig.~\ref{fig:v2-results-overview}f). Rewinding either A or B alone also reduced
performance in every replicate: the respective mean effects were 0.112 and
0.180 in coordinate control, 0.179 and 0.175 in joint control, and 0.313 and
0.337 in force control. By contrast, rewinding the protected X source actor did
not alter the matched continuation.

Both compatible source skills therefore contributed causally to the early-C
learning benefit, whereas the incompatible source did not.

\begin{table*}[!tp]
  \centering
  \caption{\textbf{Behavioral effects of module ablations.}
  Mean and sample SD (below) over eight paired runs per form and variant.
  Return A uses the first 128 return transactions, and Early C the first 192
  C transactions. Both scores range from 0 to 1; higher is better.
  Bold means mark column maxima, including ties.}
  \label{tab:v2-ablation}
  \small
  \setlength{\tabcolsep}{1pt}
  \renewcommand{\arraystretch}{1.18}
  \begin{tabular*}{\linewidth}{@{\extracolsep{\fill}}lcccccc@{}}
\toprule
 & \multicolumn{2}{c}{Coordinate} & \multicolumn{2}{c}{Three-joint} & \multicolumn{2}{c}{Hidden force} \\
\cmidrule(lr){2-3}\cmidrule(lr){4-5}\cmidrule(lr){6-7}
Variant & Return A & Early C & Return A & Early C & Return A & Early C \\
\midrule
\shortstack[l]{Shared\\context} & \shortstack{$ 0.577 $\\$\pm 0.222$} & \shortstack{$ 0.169 $\\$\pm 0.115$} & \shortstack{$ 0.660 $\\$\pm 0.163$} & \shortstack{$ 0.395 $\\$\pm 0.202$} & \shortstack{$ 0.831 $\\$\pm 0.025$} & \shortstack{$ 0.494 $\\$\pm 0.268$} \\[3pt]
\shortstack[l]{No\\composition} & \shortstack{$ \mathbf{0.864} $\\$\pm 0.023$} & \shortstack{$ 0.553 $\\$\pm 0.051$} & \shortstack{$ \mathbf{0.861} $\\$\pm 0.032$} & \shortstack{$ 0.561 $\\$\pm 0.101$} & \shortstack{$ 0.825 $\\$\pm 0.028$} & \shortstack{$ 0.484 $\\$\pm 0.062$} \\[3pt]
\shortstack[l]{No\\recurrence} & \shortstack{$ 0.855 $\\$\pm 0.026$} & \shortstack{$ 0.677 $\\$\pm 0.025$} & \shortstack{$ 0.856 $\\$\pm 0.027$} & \shortstack{$ 0.695 $\\$\pm 0.017$} & \shortstack{$ \mathbf{0.831} $\\$\pm 0.027$} & \shortstack{$ 0.712 $\\$\pm 0.030$} \\[3pt]
\midrule
\textbf{SPIRE (full)} & \shortstack{$ \mathbf{0.864} $\\$\pm 0.023$} & \shortstack{$ \mathbf{0.678} $\\$\pm 0.026$} & \shortstack{$ \mathbf{0.861} $\\$\pm 0.032$} & \shortstack{$ \mathbf{0.701} $\\$\pm 0.018$} & \shortstack{$ 0.825 $\\$\pm 0.028$} & \shortstack{$ \mathbf{0.714} $\\$\pm 0.028$} \\[3pt]
\bottomrule
\end{tabular*}

\end{table*}

\subsection{Retention and prior-learning advantage across methods}

Table~\ref{tab:v2-retention-transfer} compares immediate and late return-A
degradation across methods. Comparator function passed in 8/8
runs per method and form. Several comparators retained A in all eight runs
yet had negative mean prior-learning advantages on C. Successful retention
therefore did not ensure beneficial transfer in this evaluation.
Figure~\ref{fig:v2-results-overview}g--i
shows the per-run advantages on the unseen C composition.
For each paired replicate, we compared SPIRE's advantage with the largest
value among the five eligible task-free alternatives. SPIRE achieved the
larger value in 8/8 coordinate, 7/8 joint and 5/8 force runs, for 20/24 paired
wins and a mean gap of 0.0812. The pooled mean advantage was 0.1392 for
SPIRE and 0.0580 for the per-stream strongest eligible alternatives, giving
a ratio of means of 2.4.

Oracle EWC and SI receive phase-start information and are shown separately
from the task-free comparison. The positive learned-versus-fresh effects and
source-specific interventions distinguish SPIRE's reusable prior experience
from successful control alone.

\subsection{Source contributions to the motor drive}

We examine how protected source activity and new learning contribute to
the motor drive during C acquisition.
Figure~\ref{fig:v2-reuse-process}a compares learned-history and fresh
trajectories in each form's task space. The contribution maps show the
protected A and B source actors, \(\mathbf{u}_A\) and \(\mathbf{u}_B\), and
the active-context actor, \(\mathbf{u}_R\), on each motor-drive dimension
(Fig.~\ref{fig:v2-reuse-process}b). Once source composition is selected, the
active-context actor contributes the target residual. These components
precede clipping, exploration and recurrent event decoding; the final
control command is produced by the motor populations.

Source composition followed the initial signed confirmation probes. The
learned-history score curves rose earlier than those of the fresh controllers
(Fig.~\ref{fig:v2-reuse-process}c).

\subsection{Ablation studies}
\label{sec:v2-ablation}

We compare full SPIRE with a single shared context, disabled composition,
and removed motor recurrence (spikes and membrane state retained).
Eight new paired seed sets per form use the unchanged protocol and learning
budget (\ref{app:v2-ablation}). Table~\ref{tab:v2-ablation} separates
recovery of A from early learning of C.

\textbf{Context organization.}
Full SPIRE improves immediate return scores over the shared-context learner
by 0.287 in coordinate and 0.201 in joint control. Hidden-force return scores
are comparable, while early-C scores improve across all three forms.

\textbf{Primitive composition.}
Removing composition preserves return-A scores exactly but lowers early-C
scores by 0.126, 0.140 and 0.229 in coordinate, joint and force control.
Full SPIRE reaches the C criterion in 62.9, 61.4 and 71.6 mean transactions,
compared with 92.1, 99.6 and 138.9 without composition, isolating an
acquisition benefit beyond retained control.

\textbf{Motor recurrence.} Removal changes mean early-C scores by less than
0.007 per form, with mixed return effects: its behavioral contribution is
smaller than composition's.

\begin{samepage}
Matched 32-step motor-drive replays (\ref{app:v2-motor-interventions})
further isolate execution. Event suppression zeros commands, while recurrence
removal and private-state reset alter commands in all 24 main-evaluation
streams (Table~\ref{tab:v2-motor-interventions}). These streams emitted
3,120,222 per-neuron binary events. Replay measures command dependence,
not task-performance or energy gains.\par
\end{samepage}

\begin{table}[!t]
  \centering
  \caption{\textbf{Effects of recurrent motor interventions.}
  Mean absolute changes in normalized action commands relative to matched
  intact execution, reported as means \(\pm\) sample SD across eight SPIRE
  runs per form. Each run contributes one 32-step intervention comparison.}
  \label{tab:v2-motor-interventions}
  \small
  \footnotesize
  \setlength{\tabcolsep}{2pt}
  \renewcommand{\arraystretch}{1.12}
  \begin{tabularx}{\linewidth}{@{}>{\raggedright\arraybackslash}Xccc@{}}
\toprule
Intervention & Coordinate & Joint & Hidden force \\
\midrule
Output-event suppression & \(0.2378 \pm 0.0137\) & \(0.2133 \pm 0.0101\) & \(0.2507 \pm 0.0148\) \\
Recurrence removal & \(0.0132 \pm 0.0026\) & \(0.0129 \pm 0.0014\) & \(0.0129 \pm 0.0035\) \\
Private-state reset & \(0.0124 \pm 0.0045\) & \(0.0099 \pm 0.0008\) & \(0.0093 \pm 0.0030\) \\
\bottomrule
\end{tabularx}

\end{table}

\section{Conclusion}

SPIRE enables learned motor skills to support new behavior while preserving
recovery of earlier behavior in continuous control without task labels or
known boundaries. Action--outcome evidence guides retrieval of familiar
contexts and selection of compatible source skills for new contexts. Target
learning updates a separate residual actor while the source actors remain
protected; recurrent-spiking populations execute the resulting motor drive.

Across coordinate regulation, three-joint pose control and hidden-force
compensation, SPIRE maintained control, autonomously recovered earlier
behavior and accelerated acquisition of unseen source compositions.
Acquisition required approximately 40\% fewer mean interactions than matched
fresh controllers. Source-specific interventions established that the retained
primitives causally supported this new learning. SPIRE's mean prior-learning
advantage was 2.4 times that of the strongest eligible task-free alternative
selected for each paired stream.

Together, these findings identify complementary roles for context-organized
memory in behavior recovery and primitive composition in faster new-task
acquisition. Future work will
explore how context-guided source selection scales to larger repertoires of
learned motor skills.

\section*{Funding}
This research did not receive any specific grant from funding agencies in the
public, commercial, or not-for-profit sectors.

\appendices
\section{Implementation details}

\subsection{Controller constants}
\label{app:v2-controller}

Table~\ref{tab:v2-controller-constants} lists the coefficients shared by all
three SPIRE forms. Observation, action and transition-feature dimensions follow
Table~\ref{tab:v2-forms}; the latent width is 12 and the motor-drive width equals
the form-specific action width.

\begin{table}[!ht]
  \centering
  \caption{\textbf{Form-specific sensorimotor interface dimensions.}}
  \label{tab:v2-forms}
  \small
  \setlength{\tabcolsep}{2pt}
  \begin{tabularx}{\linewidth}{@{}Xccc@{}}
    \toprule
    Form & Observation & Action & \shortstack{Transition\\features} \\
    \midrule
    Coordinate target regulation & 4 & 2 & 21 \\
    Three-joint pose control & 8 & 3 & 39 \\
    Hidden-force compensation & 4 & 2 & 21 \\
    \bottomrule
  \end{tabularx}
\end{table}

\begin{table}[!ht]
  \centering
  \caption{\textbf{SPIRE constants shared across all three forms.}}
  \label{tab:v2-controller-constants}
  \small
  \begin{tabularx}{\linewidth}{@{}>{\raggedright\arraybackslash}Xr@{}}
    \toprule
    Quantity & Value \\
    \midrule
    Latent-context capacity & 6 \\
    Context-activity dimensions / width & 16 / 0.70 \\
    Actor / critic learning rate & 0.08 / 0.16 \\
    Utility / forward learning rate & 0.24 / 0.20 \\
    Acquisition / stable exploration & 0.25 / 0.08 \\
    Protected-source-actor expression / target-residual gain & 1.0 / 0.30 \\
    Context maturation updates & 64 \\
    Context-stability minimum / stable streak & 448 / 20 \\
    Context-stability error / outcome limits & 0.24 / 0.04 \\
    Signature samples / amplitude & 24 / 0.45 \\
    Signature match distance & 0.024 \\
    Composition cosine / margin & 0.75 / 0.20 \\
    Motor neurons per motor-drive dimension & 7 \\
    Clockwise / counterclockwise weight & 0.50 / 0.50 \\
    SNN substeps & 12 \\
    Membrane leak / recurrent decay & 0.70 / 0.45 \\
    Recurrent gain / event threshold & 0.03 / 0.78 \\
    \bottomrule
  \end{tabularx}
\end{table}

Actor, critic and forward weights are clipped to \([-3,3]\); utility weights are
clipped to \([-32,32]\). The input representations are specified in
\ref{app:v2-representations}. The signed motor codebook contains every
normalized combination in \(\{-1,0,1\}^{d_e}\) plus the hold action; the
three interfaces use \(d_e=2\) for coordinate and force control and
\(d_e=3\) for joint control.

Parameter counts are 36,936 for coordinate and force SPIRE instances and
203,226 for the joint instance. Including persistent controller state gives
38,289 and 205,385 stored scalars, respectively. These are descriptive
implementation properties.

\subsection{Sensorimotor encoding and transition features}
\label{app:v2-representations}

The sensorimotor input is
\(x_t=[o_t,a_{t-1},r_{t-1}]\in[-1,1]^{d_o+d_a+1}\). A fixed deterministic
projection \(P\) produces \(\widetilde z_t=\tanh(Px_t)\), and the encoder
preserves the leading normalized input coordinates directly:
\begin{equation}
  z_{t,j}=\begin{cases}
    x_{t,j}, & j\leq m,\\
    \widetilde z_{t,j}, & j>m,
  \end{cases}
  \qquad m=\min(d_z,\dim x_t).
  \label{eq:v2-latent}
\end{equation}
The latent width is \(d_z=12\). Define
\(g_t=[1,z_t^+,z_t^-,z_t^2]\), where \(z_t^+\) and \(z_t^-\) are the
positive and negative parts of \(z_t\), and the square is element-wise.
The first 16 entries form the normalized context-activity basis
\begin{equation}
  h_{t,j}=\frac{g_{t,j}}{\sum_{\ell=1}^{16}g_{t,\ell}},
  \qquad j=1,\ldots,16.
  \label{eq:v2-context}
\end{equation}
This 16-dimensional activity basis represents the input to the predictive
models; the six latent-context slots index the persistent models compared by
context inference. The actors and critic use the policy feature
\(f_t=[1,x_t]=[1,o_t,a_{t-1},r_{t-1}]\).

After step \(t-1\), the returned observation determines the transition-feature
vector
\begin{equation}
  \begin{aligned}
  \phi_{t-1}=\operatorname{clip}\!\bigl(
    &[1,o_t,o_t^2,o_t-o_{t-1},\\
    &o_{t-1}^2-o_t^2,a_{t-1},a_{t-1}^2],-1,1\bigr).
  \end{aligned}
  \label{eq:v2-consequence}
\end{equation}
It combines the observed state change and executed action and belongs to
\([-1,1]^{d_c}\), where \(d_c=1+4d_o+2d_a\). Squares and clipping are
element-wise. The same construction yields \(\phi_t\) after action \(a_t\).

\subsection{Auxiliary-model updates and stability}
\label{app:v2-auxiliary-learning}

The action token records the originating latent-context slot \(i\), policy
feature \(f_t\), encoding \(h_t\), motor-drive code \(k_t\) and sampled
perturbation \(\epsilon_t\). Returned feedback \((o_{t+1},r_t)\) updates
that same slot, including when the new evidence requests context confirmation.
The critic error is \(\delta^V_t=r_t-v_i^{\mathsf T}f_t\).

The actor uses Eq.~\eqref{eq:v2-actor-update} with a bounded signed perturbation
\(\epsilon_t\in\{-\sigma_t,+\sigma_t\}^{d_a}\). During acquisition,
\(\sigma_t=\sigma_0=0.25\) and \(\Gamma_{i,t}=\sigma_0^2\). Once the
context assignment is stable, physical exploration falls to \(\sigma_t=0.08\),
while
\begin{equation}
  \Gamma_{i,t}=\sigma_t\sigma_0.
  \label{eq:v2-stable-eligibility}
\end{equation}
The executed perturbation therefore becomes smaller without increasing the
actor update gain above its acquisition scale. The critic follows
\begin{equation}
  \Delta v_i=\eta_V\,
  \frac{\delta^V_t f_t}{0.1+\lVert f_t\rVert^2},
  \qquad \eta_V=0.16.
  \label{eq:v2-critic-update}
\end{equation}

The utility model learns the scalar outcome associated with the realized
transition-feature vector:
\begin{equation}
  \widehat r_t=w_i^{\mathsf T}\phi_t,
  \qquad
  \Delta w_i=\eta_U
  \frac{(r_t-\widehat r_t)\phi_t}{0.1+\lVert\phi_t\rVert^2},
  \qquad \eta_U=0.24.
  \label{eq:v2-utility-update}
\end{equation}
The action-conditioned forward model predicts the transition-feature vector
using the saved encoding,
\begin{equation}
  \widehat\phi_{i,k_t,t}=h_t^{\mathsf T}F_{i,k_t},
  \label{eq:v2-forward}
\end{equation}
and receives
\begin{equation}
  \Delta F_{i,k_t}=\eta_F
  \frac{h_t(\phi_t-\widehat\phi_{i,k_t,t})^{\mathsf T}}
  {0.03+\lVert h_t\rVert^2},
  \qquad \eta_F=0.20.
  \label{eq:v2-forward-update}
\end{equation}
These updates are applied immediately to the slot saved in the action token;
all other slots retain their state on that transaction. The protected utility
and forward templates \(\bar w_i\) and \(\bar F_{i,k}\) retain their
captured weights while the corresponding context-specific models continue
learning. Actor, critic, utility and forward-model learning is paused during
the change-detection probes.

Context maturity and context stability are separate. Maturity requires 64
local updates and permits a protected source actor to be copied from the
context-specific actor. Subsequent updates by that same slot refresh the source
copy; other slots can only read it. Context stability requires at least 448
local updates followed by 20 consecutive transactions in which the combined
utility and forward prediction error is at most 0.24 and both the current
outcome and the context-specific outcome exponential moving average exceed
0.04. Stability controls when the motor probes may run and sets the
exploration scale from context-specific experience and prediction quality.

\subsection{Controlled-system equations and evaluation thresholds}
\label{app:v2-thresholds}

For coordinate control, the environment follows
\begin{align}
  \dot p_{t+1} &= 0.48\dot p_t+0.28a_t-0.06p_t+\xi_t,\\
  p_{t+1} &= \operatorname{clip}(p_t+0.12\dot p_{t+1}),
\end{align}
with target vectors \(A=(0.82,0)\), \(B=(0,0.82)\),
\(X=(-0.82,0)\) and \(C=(0.82,0.82)\). Function requires late score at least
0.66 and late success at least 0.40.

For joint control,
\begin{align}
  \dot q_{t+1} &= 0.52\dot q_t+0.27a_t-0.08q_t+\xi_t,\\
  q_{t+1} &= \operatorname{clip}(q_t+0.10\dot q_{t+1}),
\end{align}
with \(A=(0.82,0,0)\), \(B=(0,0.82,0)\),
\(X=(0,0,-0.82)\) and \(C=(0.82,0.82,0)\). Endpoint coordinates are computed
from link lengths 0.55, 0.42 and 0.28. Function requires late score at least
0.60 and late success at least 0.55.

For hidden-force compensation,
\begin{align}
  \dot p_{t+1} &=0.86\dot p_t+0.18(a_t+b_s)-0.10p_t+\xi_t,\\
  p_{t+1} &=\operatorname{clip}(p_t+0.11\dot p_{t+1}),
\end{align}
where \(b_A=(0.46,0)\), \(b_B=(0,0.46)\),
\(b_X=(-0.46,0)\) and \(b_C=(0.46,0.46)\). Function requires late score at
least 0.70 and late success at least 0.62. Gaussian environment noise has standard
deviation 0.0025 for coordinate and force and 0.002 for joint control.

Each phase contains 640 transactions. Shared measurement windows and the
retention, prior-learning and source-reuse criteria are defined in
Section~\ref{sec:v2-measures}.

\subsection{Baseline settings}
\label{app:v2-baselines}

The cited recurrent-RL and replay methods motivate shared-protocol adaptations
\citep{ni2022,rolnick2019}: the naive ANN is an online actor--critic, while
Replay stores policy gradients rather than CLEAR's transition replay and
behavioral cloning. EWC and SI retain their parameter-protection principles
with detector- or oracle-triggered consolidation.

All recurrent ANN baselines use 128 hidden units, learning rate 0.045, critic
rate 0.12 and exploration standard deviation 0.22. EWC strength is 0.05. SI
strength and damping are 0.0003 and 0.50. The task-free change detector compares
outcome averages with rates 0.20 and 0.01, a 0.30 change threshold and an
eight-step streak. Replay stores at most 256 policy gradients and adds one
sampled past gradient at scale 0.10 per update.

Gradient-MOSAIC is a clean-room implementation based on the published
forward--inverse model equations \citep{wolpert1998mosaic,haruno2001mosaic}.
It uses four modules, 128 recurrent hidden units, actor, critic and
forward rates 0.045, 0.12 and 0.08, exploration 0.22, responsibility variance
0.03, probability floor \(10^{-5}\) and prior inertia 0.85. Gaussian
forward-prediction likelihood and previous responsibility determine the next
posterior; that posterior weights both forward and inverse-policy updates.

\subsection{Replication and information access}
\label{app:v2-replication}

The evaluation contains eight replicates for every form--method cell. Methods
within a form and replicate share the environment seed. Fresh-C controls use the same
controller seed as their learned counterpart and align random state at C
entry.

SPIRE, naive ANN, task-free EWC, task-free SI, replay and gradient-MOSAIC do
not receive task labels, boundaries or phase information. Oracle EWC and SI
receive phase-start information solely to trigger consolidation. No controller receives target vectors,
hidden force vectors, evaluator success, threshold values or future state.

Each SPIRE run in the main evaluation contains 3,200 main-stream transactions, 640 fresh-C
transactions and six matched C continuations after endogenous composition
opens: intact, sham, and four source-rewind conditions.

\subsection{Recurrent motor interventions}
\label{app:v2-motor-interventions}

For each SPIRE run, 32 recorded motor drives are replayed from matched
checkpoints under intact execution, event suppression, recurrence removal
or private-state reset. The run-level effect in
Table~\ref{tab:v2-motor-interventions} averages absolute action differences
from intact execution over steps and action dimensions.

\subsection{Module-ablation protocol}
\label{app:v2-ablation}

The 96 runs pair controller/environment seeds across variants, with 3,200
stream and 640 fresh-C transactions each. Shared context disables inference,
probes and retrieval, leaving one online actor and consequently removing
distinct source storage. No-composition retains context inference/storage
and uses standalone actor gain 1. Recurrence removal includes probes.
Unreached C thresholds remain missing (shared context: 8/8 coordinate,
3/8 joint); all other cells reach the criterion.

\printbibliography

\end{document}